\documentclass{article}

\usepackage{times}
\usepackage{graphicx} 
\usepackage{subfigure}

\usepackage{natbib}

\usepackage{algorithm}
\usepackage{algorithmic}

\usepackage{hyperref}

\usepackage[accepted]{icml2016}

\usepackage{amsmath}

\newcommand{\be}{\begin{equation}}
\newcommand{\ee}{\end{equation}}
\newcommand{\bea}{\begin{eqnarray}}
\newcommand{\eea}{\end{eqnarray}}

\newcommand{\BEAS}{\begin{eqnarray*}}
\newcommand{\EEAS}{\end{eqnarray*}}
\newcommand{\BEA}{\begin{eqnarray}}
\newcommand{\EEA}{\end{eqnarray}}
\newcommand{\BEQ}{\begin{equation}}
\newcommand{\EEQ}{\end{equation}}
\newcommand{\BEQS}{\begin{equation*}}
\newcommand{\EEQS}{\end{equation*}}
\newcommand{\BIT}{\begin{itemize}}
\newcommand{\EIT}{\end{itemize}}
\newcommand{\BNUM}{\begin{enumerate}}
\newcommand{\ENUM}{\end{enumerate}}
\newcommand{\BP}{\begin{proof}}
\newcommand{\EP}{\end{proof}}

\newcommand{\BA}{\begin{array}}
\newcommand{\EA}{\end{array}}
\newcommand{\BT}{\begin{tabular}}
\newcommand{\ET}{\end{tabular}}

\icmltitlerunning{Overcoming Challenges in Fixed Point Training of Deep Convolutional Networks}

\begin{document}

\twocolumn[
\icmltitle{Overcoming Challenges in Fixed Point Training of Deep Convolutional Networks}

\icmlauthor{Darryl D. Lin}{darryl.dlin@gmail.com}
\icmladdress{Qualcomm Research, San Diego, CA 92121 USA}
\icmlauthor{Sachin S. Talathi}{talathi@gmail.com}
\icmladdress{Qualcomm Research, San Diego, CA 92121 USA}

\icmlkeywords{deep learning, quantization, fixed point}

\vskip 0.3in
]

\begin{abstract}
It is known that training deep neural networks, in particular, deep convolutional networks, with aggressively reduced numerical precision is challenging. The stochastic gradient descent algorithm becomes unstable in the presence of noisy gradient updates resulting from arithmetic with limited numeric precision. One of the well-accepted solutions facilitating the training of low precision fixed point networks is stochastic rounding. However, to the best of our knowledge, the source of the instability in training neural networks with noisy gradient updates has not been well investigated. This work is an attempt to draw a theoretical connection between low numerical precision and training algorithm stability. In doing so, we will also propose and verify through experiments methods that are able to improve the training performance of deep convolutional networks in fixed point.
\end{abstract}

\section{Introduction}

Deep convolutional networks (DCNs) have demonstrated state-of-the-art performance in many machine learning tasks such as image classification \citep{Krizhevsky_2012} and speech recognition \citep{Deng_2013}. However, the complexity and the size of DCNs have limited their use in mobile applications and embedded systems. One reason is related to the hit on performance (in terms of accuracy on a given machine learning task) that these networks take when they are deployed with data representations using reduced numeric precision. A potential avenue to alleviate this problem is to fine-tune pre-trained floating point DCNs using data representations with reduced numeric precision. However, the training algorithms have a strong tendency to diverge when the precision of network parameters and features are too low \citep{han2015deep, courbariaux2014low}.

More recently, several works have touched upon the issue of training deep networks with low numerical precision \citep{gupta2015deep, Zhouhanlin_2014, gysel2016}. In all of these works  \emph{stochastic rounding} has been the key to improving the convergence properties of the training algorithm, which in turn has enabled training of deep networks with relatively small bit-widths.

However, to the best of our knowledge, there is a limited understanding from a theoretical point of view as to why low precision networks lead to training difficulties. In this paper, we attempt offer a theoretical insight into the root cause of the numerical instability when training DCNs with limited numeric precision representations. In doing so, we will also propose a few solutions to combat such instability in order to improve the training outcome. These proposals are not meant to replace stochastic rounding. Rather, they are complementary techniques. To clearly demonstrate the effectiveness of our proposed solutions, we will not perform stochastic rounding in the experiments. We intend to combine stochastic rounding and our proposed solutions in future works.

This work will focus on fine-tuning a pre-trained floating point DCN in fixed point. While most of the analysis apply also to the case of training a fixed point network from scratch, some discussions may be applicable to the fixed point fine-tuning scenario alone.

\section{Low Precision and Back-Propagation} \label{sec:back_prop}

In this section, we will investigate the origin of instability in the network training phase when low precision weights and activations are used. The outcome of this effort will shed light on possible avenues to alleviate the problem.

\subsection{Effective Activation Function} \label{sec:effective_act}

The computation of activations in the forward pass of a deep network can be written as:
\BEQ \label{eq:activation}
	a_i^{(l)} = \sum_j w_{i,j}^{(l)} \cdot g(a_j^{(l-1)}),
\EEQ
where $a_i^{(l)}$ denotes the $i$-th activation in the $l$-th layer, $w_{i,j}^{(l)}$ represents the $(i,j)$-th weight value in the $l$-th layer. And $g(\cdot)$ is the activation function.

Note that here we assume both the activations and weights are full precision values. Now consider the case where only the weights are low precision fixed point values. From the forward pass perspective, (\ref{eq:activation}) still holds.

\begin{figure}[htb]
\begin{center}
    \includegraphics[width=1.0\linewidth, bb=45 670 505 785]{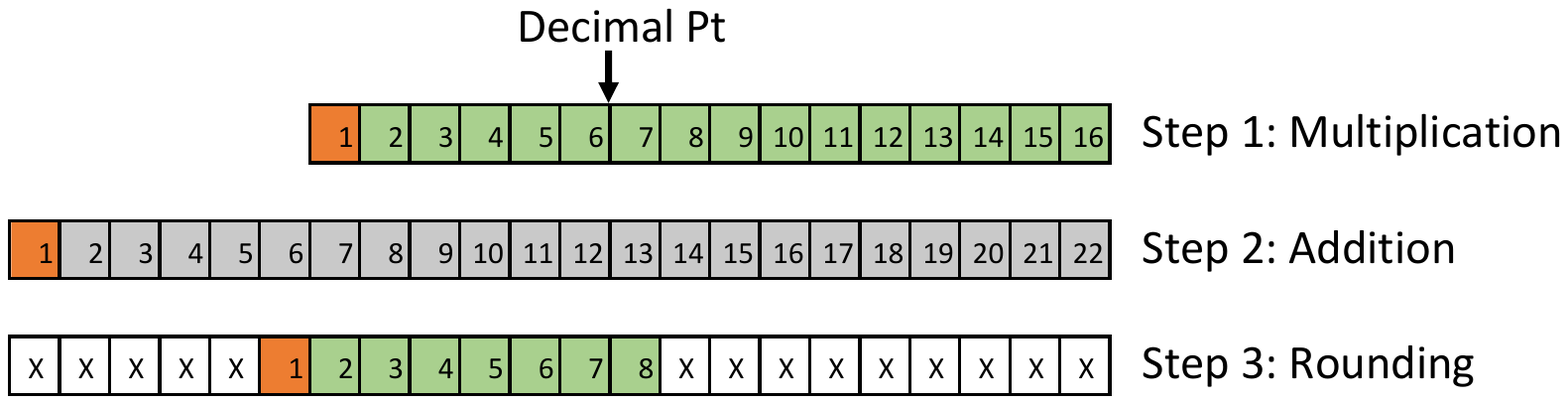}
\caption{Evaluation of activation as quantization}\label{fig:quantize_sum}
\end{center}
\end{figure}

However, when we introduce low precision activations into the equation, (\ref{eq:activation}) is no longer an accurate description of how the activations propagate. To see this, we may consider the evaluation of $a_i^{(l)}$ in fixed point representation as in Figure \ref{fig:quantize_sum}.

In Figure \ref{fig:quantize_sum}, three operations are depicted:
\BIT
\item Step 1: Compute $w\cdot g(a)$. Assuming both $w$ and $g(a)$ are 8-bit fixed point values, the product is a 16-bit value.
\item Step 2: Compute $\sum w\cdot g(a)$. The size of the accumulator is larger than 16-bit to prevent overflow.
\item Step 3: The outcome of $\sum w\cdot g(a)$ is rounded and truncated to produce an 8-bit activation value.
\EIT

\begin{figure}[htb]
\begin{center}
    \includegraphics[width=1.0\linewidth, bb=155 610 430 710]{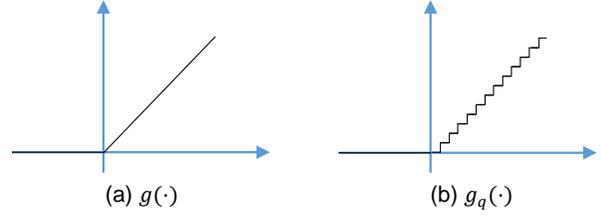}
\caption{The presumed and actual ReLU function in low precision networks}\label{fig:f_x}
\end{center}
\end{figure}

Step 3 is a quantization step that reduces the precision of the value calculated based on (\ref{eq:activation}) in keeping with the desired fixed point precision of layer $l$. In essence, assuming ReLU, the effective activation function experienced by the features in the network is as shown in Figure \ref{fig:f_x}(b), rather than \ref{fig:f_x}(a).

\subsection{Gradient Mismatch}

In back-propagation, denoting the cost function as $C$, an important equation that dictates how the ``error signal'', $\frac{\partial C}{\partial a_i^{(l)}}$, propagates down the network is expressed as follows:
\BEQ \label{eq:derivative_x}
	\frac{\partial C}{\partial a_i^{(l)}} = g'(a_i^{(l)})\cdot \sum_j w_{i,j}^{(l+1)} \frac{\partial C}{\partial a_j^{(l+1)}}.
\EEQ

The value of $-\frac{\partial C}{\partial a_i^{(l)}}$ indicates the direction in which $a_i^{(l)}$ should move in order to improve the cost function. Playing a crucial role in (\ref{eq:derivative_x}) is the derivative of the activation function, $g'(a_i^{(l)})$. In a software environment that implements SGD, original activation functions in the form of Figure \ref{fig:f_x}(a) is assumed. However, as explained in Section \ref{sec:effective_act}, the effective activation function in a fixed point network is a non-differentiable function as described in Figure \ref{fig:f_x}(b).

This disagreement between the presumed and the actual activation function is the origin of what we call the ``gradient mismatch'' problem. When the bit-widths of the weights and activations are large, the gradient of the original activation function offers a good approximation to that of the quantized activation function. However, the mismatch will start to impact the stability of SGD when the bit-widths become too small (step sizes become too large).

The gradient mismatch problem also exacerbates as the error signal propagates deeper down the network, because every time the presumed $g'(a_i^{(l)})$ is used, additional errors are introduced in the gradient computation. Since the gradients w.r.t. the weights are directly based on the gradients w.r.t. the activations,
\BEQ
	\frac{\partial C}{\partial w_{i,j}^{(l)}} = g(a_i^{(l-1)})\cdot \frac{\partial C}{\partial a_j^{(l)}},
\EEQ
the weight updates become increasingly inaccurate as the error propagates into lower layers of the network. Hence training networks in fixed point is much more challenging in deeper networks than in shallower networks.

\subsection{Potential Solutions}

Having understood the source of the issue, we will propose a few methods to help overcome the challenges of training or fine-tuning a fixed point network. The obvious approach of replacing the perceived activation function with the effective activation function that takes quantization into account is not viable because the effective activation function is not differentiable. However, some alternatives may help improve convergence during model training to avoid the gradient mismatch problem.

\subsubsection{Proposal 1: Low Precision Weights and Full Precision Activations}
Recognizing that the main obstacle of training in fixed point is the low precision activations, we may train a network with the desired precision for the weights, while keeping the activations floating point or with relatively high precision. After training, the network can be adapted to run with lower precision activations.

\subsubsection{Proposal 2: Fine-Tuning Top Layer(s) Only}
As the analysis in Section \ref{sec:back_prop} shows, when the activation precision is low, weight updates of top layers are more reliable than lower layers, because the gradient mismatch builds up from the top of the network to the bottom. Therefore, while it may not be possible to fine-tune the entire network, it may be possible to fine-tune only the top layers without incurring convergence issues.

\subsubsection{Proposal 3: Bottom-to-Top Iterative Fine-Tuning}
The bottom-to-top iterative fine-tuning scheme is a training algorithm designed to avoid gradient mismatch. At the same time, it allows the entire network to be fine-tuned. For example, consider a network with 4 layers. Table \ref{tab:finetune_phases} offers an illustration of how fine-tuning is divided into phases where one layer is fine-tuned in each phase.

\begin{table}[h]
\caption{Example showing the phases of iterative fine-tuning}
\small\label{tab:finetune_phases}
\begin{center}
\begin{tabular}{|c|cc|cc|cc|cc}
\hline
\multicolumn{1}{|c|}{}  &\multicolumn{2}{c|}{Phase 1} &\multicolumn{2}{c|}{Phase 2} &\multicolumn{2}{c|}{Phase 3}\\
\hline
&\parbox[t]{0.4cm}{\centering Acts} &\parbox[t]{0.4cm}{\centering Wgts} &\parbox[t]{0.4cm}{\centering Acts} &\parbox[t]{0.4cm}{\centering Wgts} &\parbox[t]{0.4cm}{\centering Acts} &\parbox[t]{0.4cm}{\centering Wgts}\\
\hline
\hline
Layer4       &Float	&-	     &Float	&-	     &Float	&update	 \\
Layer3       &Float	&-	     &Float	&update	 &FixPt	&-	     \\
Layer2       &Float	&update	 &FixPt	&-	     &FixPt	&-	     \\
Layer1       &FixPt	&-	     &FixPt	&-	     &FixPt	&-	     \\
\hline
\end{tabular}
\end{center}
\end{table}

Each phase of fine-tuning, consisting of 1 or multiple epochs, updates the weights of one of the layers (weights can follow the desired fixed point format without special treatment). As shown in Table \ref{tab:finetune_phases}, Phase 1 fine-tunes the weights of Layer2. After Phase 1 is complete, Phase 2 fine-tunes the weights of Layer3 while keeping the weights of all other layers static. Then Phase 3 fine-tunes Layer4 in a similar manner. Note that Layer1 weights are quantized but never fine-tuned.

Also of importance is how the number format of activations change over the phases. Initially during Phase 1, only the bottom layer (Layer1) activations are in fixed point, but in Phase 2, both Layer1 and Layer2 activations are in fixed point. In the last phase of fine-tuning, only the output of the final layer remains floating point. All other activations have been turned into fixed point. The gradual turning on of fixed point activations is designed to prevent gradient mismatch completely. Careful inspection of the algorithm shows that, whenever the weights of a particular layer are updated, the gradients are always back-propagated from layers with only floating point activations.

\section{Experiments}

In this section, we examine the effectiveness of the proposed solutions based on a deep convolutional network we developed for the ImageNet classification task \footnote{Proprietary Information, Qualcomm Inc }. The network has 12 convolutional layers and 5 fully-connected layers. We choose this network to experiment because, as we have shown in a network designed for CIFAR-10 classification \citep{Lin_2016}, fine-tuning a relatively shallow fixed point network does not pose convergence challenges even when the bit-widths are small.

\begin{table}[htb]
\caption{ImageNet classification Top-5 error rate (\%): No fine-tuning}
\small
\label{tab:no_finetune}
\begin{center}
\begin{tabular}{|c|cccc|}
\hline
\multicolumn{1}{|c|}{Activation}  &\multicolumn{4}{c|}{Weight Bit-width} \\
\multicolumn{1}{|c|}{Bit-width} &\multicolumn{1}{c}{4}  &\multicolumn{1}{c}{8}  &\multicolumn{1}{c}{16}  &\multicolumn{1}{c|}{Float}\\
\hline
\hline
4       &98.6  &33.4  &32.9  &32.7\\
8       &97.1  &19.3  &18.0  &18.2\\
16      &96.6  &15.0  &14.3  &14.4\\
Float   &96.6  &14.1  &13.9  &13.8\\
\hline
\end{tabular}
\end{center}
\end{table}

The baseline for the experiment is the DCN network that is quantized based on the algorithm presented in \citet{Lin_2016} without fine-tuning. The Top-5 error rates of these networks, for different weight and activation bit-width combinations, are listed in Table \ref{tab:no_finetune}. Note that for all the fixed point experiments in this paper, the output activations of the final fully-connected layer is always set to a bit-width of 16. We do not try to reduce the precision of this quantity because the subsequent softmax layer is rather sensitive to low precision inputs and it is an insignificant overhead to the network overall.

\begin{table}[htb]
\caption{ImageNet classification Top-5 error rate (\%): Plain vanilla fine-tuning (``n/a''=``fails to converge'')}
\small
\label{tab:vanilla_finetune}
\begin{center}
\begin{tabular}{|c|cccc|}
\hline
\multicolumn{1}{|c|}{Activation}  &\multicolumn{4}{c|}{Weight Bit-width} \\
\multicolumn{1}{|c|}{Bit-width} &\multicolumn{1}{c}{4}  &\multicolumn{1}{c}{8}  &\multicolumn{1}{c}{16}  &\multicolumn{1}{c|}{Float}\\
\hline
\hline
4       &n/a  &n/a  &n/a  &n/a\\
8       &n/a  &19.3  &n/a  &n/a\\
16      &21.0  &n/a  &n/a  &n/a\\
Float   &22.2  &13.5  &13.3  &13.8\\
\hline
\end{tabular}
\end{center}
\end{table}

To further improve the accuracy beyond Table \ref{tab:no_finetune}, we perform fine-tuning on these networks subject to the corresponding fixed point bit-width constraints of the weights and activations. Table \ref{tab:vanilla_finetune} shows that, while fine-tuning improves some scenarios (for example, 16-bit activations and 4-bit weights), it fails to converge for most of the settings where the activations are in fixed point. This interesting observation validates the analysis in Section \ref{sec:back_prop} showing that the stability problem is due to the low precision of activations, not weights. We note that for these and all the subsequent fine-tuning experiments, we did not perform any hyperparameter optimization of the training parameters and it is quite possible to identify a set of training hyperparameters for which the quantized network may train successfully. 

\subsection{Proposal 1}

\begin{table}[htb]
\caption{ImageNet classification Top-5 error rate (\%): Use fixed point activations in networks trained with floating point activations (Proposal 1)}
\small
\label{tab:adopt_float}
\begin{center}
\begin{tabular}{|c|cccc|}
\hline
\multicolumn{1}{|c|}{Activation}  &\multicolumn{4}{c|}{Weight Bit-width} \\
\multicolumn{1}{|c|}{Bit-width} &\multicolumn{1}{c}{4}  &\multicolumn{1}{c}{8}  &\multicolumn{1}{c}{16}  &\multicolumn{1}{c|}{Float}\\
\hline
\hline
4       &45.6  &32.0  &31.3  &32.7\\
8       &25.1  &16.8  &16.8  &18.2\\
16      &22.5  &13.9  &13.8  &14.4\\
Float   &22.2  &13.5  &13.3  &13.8\\
\hline
\end{tabular}
\end{center}
\end{table}

The networks on the last row of Table \ref{tab:vanilla_finetune} are already trained with the desired weight precision. We can directly use them to run with different activation precision. Table \ref{tab:adopt_float} lists the classification accuracy of this approach. It is seen that we can achieve fairly good classification accuracy for different activation bit-widths.

\subsection{Proposal 2}

Using the networks on the last row of Table \ref{tab:vanilla_finetune} as the baseline, we can continue to fine-tune only the weights of the top few layers. It is possible to fine-tune the top layers because the effect of gradient mismatch accumulates toward the lower layers of the network, but the impact on the top layers is relatively small.

\begin{table}[htb]
\caption{ImageNet classification Top-5 error rate (\%): Fine-tune the top fully-connected layer (Proposal 2)}
\small
\label{tab:finetune_top_layer}
\begin{center}
\begin{tabular}{|c|cccc|}
\hline
\multicolumn{1}{|c|}{Activation}  &\multicolumn{4}{c|}{Weight Bit-width} \\
\multicolumn{1}{|c|}{Bit-width} &\multicolumn{1}{c}{4}  &\multicolumn{1}{c}{8}  &\multicolumn{1}{c}{16}  &\multicolumn{1}{c|}{Float}\\
\hline
\hline
4       &37.1  &23.8  &23.3  &23.5\\
8       &22.8  &15.6  &15.7  &16.2\\
16      &21.2  &13.7  &13.5  &13.7\\
Float   &22.2  &13.5  &13.3  &13.8\\
\hline
\end{tabular}
\end{center}
\end{table}

Table \ref{tab:finetune_top_layer} demonstrates the results of fine-tuning only the top fully-connected layer in the network. It is seen that fine-tuning the top layer offers a small boost in accuracy compared to the networks in Table \ref{tab:adopt_float}.

\subsection{Proposal 3}

Again using the network on the last row of Table \ref{tab:vanilla_finetune} as the fine-tuning baseline, we iteratively fine-tune the network from the bottom to the top, one layer at a time, according to the algorithm prescribed in Table \ref{tab:finetune_phases}. This procedure ensures that each layer has accurate gradient information when the weights are updated.

\begin{table}[htb]
\caption{ImageNet classification Top-5 error rate (\%): Iterative fine-tuning from bottom layer to top layer (Proposal 3)}
\small
\label{tab:bottom_to_top}
\begin{center}
\begin{tabular}{|c|cccc|}
\hline
\multicolumn{1}{|c|}{Activation}  &\multicolumn{4}{c|}{Weight Bit-width} \\
\multicolumn{1}{|c|}{Bit-width} &\multicolumn{1}{c}{4}  &\multicolumn{1}{c}{8}  &\multicolumn{1}{c}{16}  &\multicolumn{1}{c|}{Float}\\
\hline
\hline
4       &25.3  &18.4  &18.3  &18.2\\
8       &19.3  &15.2  &14.1  &14.1\\
16      &18.8  &13.2  &13.2  &13.5\\
Float   &22.2  &13.5  &13.3  &13.8\\
\hline
\end{tabular}
\end{center}
\end{table}

As seen in Table \ref{tab:bottom_to_top}, this approach provides a significant performance boost compared to the previous solutions. Even a network with 4-bit weights and 4-bit activations is able to achieve Top-5 error rate of 25.3\%. Some of the entries in the table have better accuracy than the floating point baseline. This may be attributed to the regularization effect of the added quantization noisy during training \citep{Zhouhanlin_2014}.

\section{Conclusion}
In this paper, we studied the effect of low numerical precision of weights and activations on the accuracy of gradient computation during back-propagation. Our analysis showed that low precision weights are benign, but low precision activations have a detrimental impact on the computed gradients. The errors in gradient computation accumulate during back-propagation and may slow and even prevent the successful convergence of gradient descent when the network is sufficiently deep.

We proposed a few solutions to combat this problem and demonstrated through experiments their effectiveness on the ImageNet classification task. We plan to combine stochastic rounding and our proposed solutions in future works.

\bibliography{icml2016_conference}
\bibliographystyle{icml2016}

\end{document}